\documentclass[pdflatex,sn-mathphys-num]{sn-jnl}% Math and Physical Sciences Numbered Reference Style
\usepackage{graphicx}%
\usepackage{multirow}%
\usepackage{amsmath,amssymb,amsfonts}%
\usepackage{amsthm}%
\usepackage{mathrsfs}%
\usepackage[title]{appendix}%
\usepackage{xcolor}%
\usepackage{textcomp}%
\usepackage{manyfoot}%
\usepackage{booktabs}%
\usepackage{array}
\usepackage{algorithm}%
\usepackage{algorithmicx}%
\usepackage{algpseudocode}%
\usepackage{listings}%
\usepackage{bm}
\usepackage{wrapfig}
\theoremstyle{thmstyleone}%
\theoremstyle{thmstyletwo}%

\theoremstyle{thmstylethree}%

\begin{document}

\title[Article Title]{Reasoning Language Model for Personalized Lung Cancer Screening}


\author{\fnm{Chuang} \sur{Niu}}


\author{\fnm{Ge} \sur{Wang}}

\affil{\orgdiv{Department of Biomedical Engineering,
School of Engineering,
Center for Computational Innovations,
Center for Biotechnology \& Interdisciplinary Studies}, \orgname{Rensselaer Polytechnic Institute}, \orgaddress{\street{110 8th Street}, \city{Troy}, \postcode{12180}, \state{NY}, \country{USA}}}

% \affil[2]{\orgdiv{Department}, \orgname{Organization}, \orgaddress{\street{Street}, \city{City}, \postcode{10587}, \state{State}, \country{Country}}}

% \affil[3]{\orgdiv{Department}, \orgname{Organization}, \orgaddress{\street{Street}, \city{City}, \postcode{610101}, \state{State}, \country{Country}}}


\abstract{Accurate risk assessment in lung cancer screening is critical for enabling early cancer detection and minimizing unnecessary invasive procedures. The Lung CT Screening Reporting and Data System (Lung-RADS) has been widely used as the standard framework for patient management and follow-up. Nevertheless, Lung-RADS faces trade-offs between sensitivity and specificity, as it stratifies risk solely based on lung nodule characteristics without incorporating various risk factors. Here we propose a reasoning language model (RLM) to integrate radiology findings with longitudinal medical records for individualized lung cancer risk assessment. Through a systematic study including dataset construction and distillation, supervised fine-tuning, reinforcement learning, and comprehensive evaluation, our model makes significant improvements in risk prediction performance on datasets in the national lung screening trial. Notably, RLM can decompose the risk evaluation task into sub-components, analyze the contributions of diverse risk factors, and synthesize them into a final risk score computed using our data-driven system equation. Our approach improves both predictive accuracy and monitorability through the chain of thought reasoning process, thereby facilitating clinical translation into lung cancer screening.}

\maketitle

\section{Introduction}\label{sec1}
Lung cancer remains the single most diagnosed and deadliest form of cancer worldwide for both men and women~\cite{lung}.
According to the World Health Organization, there are about 2.2 million new lung cancer cases and 1.8 million deaths each year, roughly one-fifth of all cancer deaths, exceeding the combined fatalities from breast and colorectal cancers. The disease often presents at an advanced stage, contributing to poor overall survival (global 5-year survival ~20\%). Indeed, over half of lung cancers are diagnosed at stage IV, whereas only 15–25\% are caught early (stage I). This late detection drives the grim prognosis, as advanced lung cancer has limited curative options. In contrast, if lung cancer is detected while still localized, an estimated 68–92\% of patients survive at least five years. Such a stark difference underscores the crucial role of early detection~\cite{amicizia2023systematic}. 

Lung cancer screening (LCS) aims to identify tumors at a curable stage before symptoms arise, thereby reducing mortality.
The primary modality for LCS is low-dose computed tomography (LDCT), which uses reduced radiation dose to image the lungs in high detail and detects early-stage malignancies. Annual LDCT screening in high-risk individuals has proven capable of catching lung cancers while still localized for effective treatment and even curation. Evidence from large randomized trials unequivocally shows that LDCT screening can significantly reduce lung cancer mortality in high-risk populations. For example, the National Lung Screening Trial (NLST) in the United States~\cite{national2011reduced}, the NELSON trial in Netherlands and Belgium~\cite{de2020reduced}, and the MILD trial in Italy~\cite{pastorino2019ten} have demonstrated that LDCT LCS can reduce the lung cancer mortality by 20\%, 24\%, and 39\%, respectively.  

However, LCS faces major implementation challenges.
First, accurate risk stratification is the core challenge in evaluating the possibilities of developing lung cancer. The Lung Imaging Reporting and Data System (Lung-RADS), introduced by the American College of Radiology (ACR) in 2014, standardizes LDCT screening reports and is widely adopted as a classification and management tool for follow-up recommendations~\cite{pinsky2015performance}. Nevertheless, Lung-RADS still faces the trade-off between sensitivity and specificity. Notably, Lung-RADS is based on LDCT imaging reports without integrating personal risk factors, such as age, smoking intensity, family history, and occupational exposure. 
On the other hand, the LCS screening rate is extremely low ($<10\%$) among eligible smokers~\cite{fedewa2021state}, due to various reasons such as limitations in resource access, disparities across socioeconomic and demographic lines, and a global shortage of radiology expertise for providing LCS~\cite{jonas2021screening,rivera2016lung, triplette2022patient, lin2022patient, nunez2021adherence,glover2017socioeconomic, tseng2019relationship,wang2019barriers}.
Hence, there is an important and immediate need for multidisciplinary efforts to broadly and optimally implement personalized LCS and minimize lung cancer mortality.

The rapid advancement of artificial intelligence (AI), particularly the emergence of large language models (LLMs)~\cite{bommasani2021opportunities}, presents unprecedented opportunities to enhance the precision of risk assessment through large-scale data-driven methodologies. This potential is especially promising and benefitial in the context of LCS, as lung cancer has been the most commonly diagnosed malignancy and has accumulated extensive data over the past decades. 
A deep learning method was proposed for lung cancer detection and risk estimation with LDCT in an end-to-end manner, demonstrating that the AI model outperformed radiologists in terms of both false negative and false positive rates~\cite{ardila2019end}. Then, the Sybil model was developed for lung cancer risk prediction using a single LDCT scan, which can predict up to six year risks of developing lung cancer~\cite{mikhael2023sybil}. Most recently, a medical multimodal multitask foundation model (M3FM) was designed to integrate multimodal data and perform multiple LCS tasks. Due to the larger scale of the model and datasets, M3FM redefined the state-of-the-art (SOTA) performance in 17 LCS tasks~\cite{niu2025medical}.
While the performance of LCS models continues to improve, a major barrier to clinical translation remains how to monitor the reliability of AI predictions and establish trust.

In this study, we propose a reasoning language model (RLM) for lung cancer risk assessment by integrating LDCT imaging findings and various individual risk factors, with monitorability through its chain-of-thought (CoT) reasoning process~\cite{wei2022chain, korbak2025chain}.
Reasoning is a most critical frontiers in AI research. While early LLMs excelled at generic text generation, they often failed in tasks requiring logical consistency, multi-step inference, or domain-specific reasoning. 
RLMs aim to overcome these limitations by explicitly modeling step-by-step thinking.
From a theoretical perspective, CoT reasoning makes a Transformer strictly more powerful~\cite{li2024chain}.
Therefore, reasoning has become a core technique to improve performance of LLMs by scaling the test time compute across domains, especially for difficult tasks, such as mathematics, coding, and scientific discovery.
In practice, LLMs learn a strong “natural language prior” during pretraining such that they tend to use the CoT in a manner similar to the way that humans use natural language.
It is widely believed that current AI systems satisfy the externalized reasoning property:
``For sufficiently difficult tasks, Transformers must use CoT as a form of working memory. By default, humans can understand this chain of thought."
Thus, reasoning in human language offers a unique opportunity for AI safety by monitoring CoT with the intent of misbehaving~\cite{korbak2025chain}.

In this context, we assume that reasoning is an effective component in accurate and monitorable lung cancer risk assessment, as it is a highly sophisticated task requiring systematic analysis of various risk factors. 
Although conceptually promising, reasoning capabilities for LCS have never been built nor evaluated.
There are two primary challenges: 1) how to build a high-quality dataset for lung cancer risk assessment involving radiology findings and diverse risk factors, and 2) how to optimize the model in a scalable and stable manner to induce the CoT thinking process with monitorability in lung cancer risk assessment. 
To this end, we first present a data curation and augmentation workflow and construct a large-scale dataset from NLST.
Then, we develop the first-of-its-kind reasoning LLM capable of breaking down the LDCT imaging findings and individual medical records into a set of risk factors, analyzing the risk score of each factor, integrating and refining the final risk score by further considerations.
Extensive results show that LLM models trained with reasoning abilities consistently and significantly outperform those without reasoning and achieved significantly better results than Lung-RADS in lung cancer risk assessment.
Importantly, our experiments show that the CoT thinking process offers a way to monitor the reliability of LLM, which is important not only for inspecting the training process but also for clinical translation.

\section{Approach}\label{sec2}

\subsection{Problem Formulation}
In this study, the task of lung cancer risk assessment is to predict a risk score within a certain time frame, given the longitudinal LDCT imaging reports, patient demographics, and the history of smoking , disease, personal cancer, family lung cancer, work, and alcohol. All individual information is described in a free text format and then converted into a sequence of tokens, $\bm{x}_{1:N} = [x_1, x_2, \cdots, x_N]$, $N$ is the number of input tokens, and then the LLM generates a sequence of output tokens including reasoning tokens $\bm{r}_{1:T}$ enclosed between special tokens $<think>$ and $</think>$ and answer tokens $\bm{y}_{1:K}$ that contains the risk score $s\in [0, 1]$ in the latex format $\boxed{s}$, where $T$ and $K$ are the numbers of reasoning and answering tokens, respectively. The modeling process can be formulated as a probability distribution over sequences of tokens:
\begin{align}
    &P_\theta(\bm{r}, \bm{y} \mid \bm{x}) = P_\theta(\bm{r} \mid \bm{x}) \, P_\theta(\bm{y} \mid \bm{x}, \bm{r}) \\
    &= \prod_{t=1}^{K} P_\theta(\bm{r}_k \mid \bm{x}, \bm{r}_{<k}) \prod_{t=1}^{T} P_\theta(\bm{y}_t \mid \bm{x}, \bm{r}, \bm{y}_{<t}),
\end{align}
where $\theta$ denotes the model parameters. We expect a score $s$ inside the answer part, which can be deterministically extracted as:
\begin{equation}
    s = g(\bm{y}), \quad g: \text{string} \to \mathbb{R}
\end{equation}
where $g$ is a deterministic parser that finds the risk score $s$ in the output text. When the score $s$ cannot be extracted in the predefined format, it will be considered a wrong prediction.
Moreover, the answer is expected to break down the provided imaging findings and clinical data into a set of risk factors, analyze the risk score of each factor, integrate and refine the final risk score by further considerations.

\subsection{Dataset Construction}
Our current datasets were constructed with the data collected in NLST, which is a randomized trial for evaluating LCS with 3D LDCT versus 2D chest radiography, demonstrating that screening with LDCT lowered lung cancer mortality by 20\%. The 26,722 participants in the LDCT screening arm were enrolled from August 2002 through April 2004 in 33 medical institutions. Participants underwent three screenings at 1-year intervals from August 2002 through September 2007. The follow-up data were collected until December 31, 2009. During the whole process, diverse data were recorded, including demographics, smoking history, disease history, multiple CT series, key abnormal findings by radiologists in fully structured reports, pathology test results for lung cancer, follow-up data, and vital status.
The key LDCT imaging findings in NLST consist of abnormalities of lung nodules including the presence of lung nodules and their location, size, margin, and attenuation properties, and opportunistic abnormalities including atelectasis, pleural thickening/effusion, non-calcied hilar/mediastinal adenopathy/mass, chest wall abnormality (bone destruction, metastasis, etc.), consolidation, emphysema, reticular/reticulonodular opacities/honeycombing/brosis/scar, cardiovascular disease, and their changes relative previous findings.

\begin{wraptable}{r}{0.4\textwidth} % r = right, l = left
  \centering
  \caption{Patient-Centric Data Elements.}
  \includegraphics[width=0.38\textwidth,page=1]{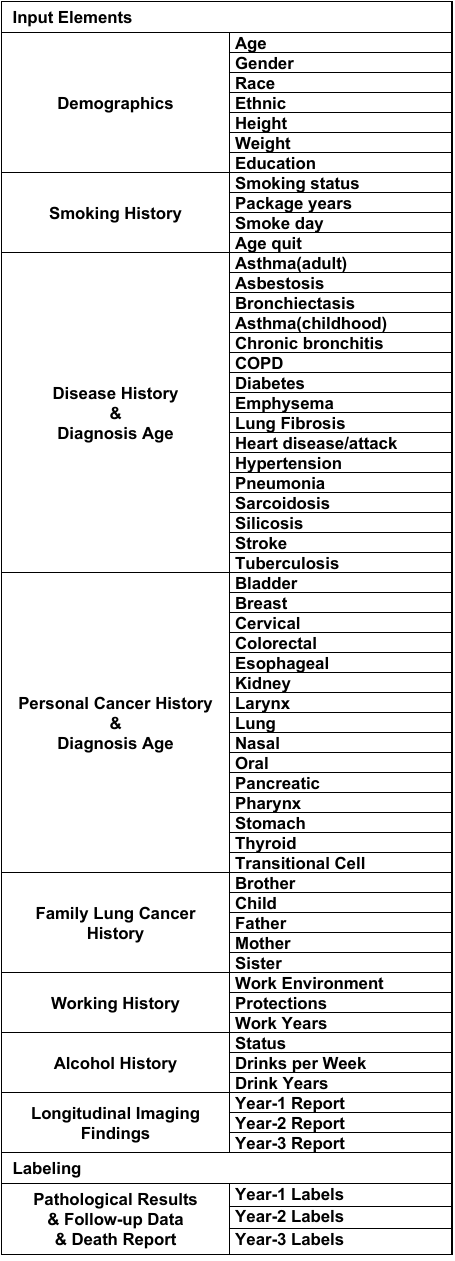}
  \label{tab:data}
\end{wraptable}

During the curation of datasets, we aligned all clinical elements in a patient-centric manner, as shown in Table~\ref{tab:data}.
The input data contain individual risk factors including demographics, smoking history, disease history, personal cancer history, family lung cancer history, work history, and alcohol history, collected at the beginning of enrollment in NLST, and longitudinal LDCT imaging findings up to three years.
At each screening year, we input the imaging findings from the current and previous years if available.
Correspondingly, we calculated the ground-truth labels at each screening year as: 1) The lung cancer risk score is 0 if no pathology confirmed lung cancer within $n$ years and the follow-up years is larger than or equal to $n$ years and the patient did not die of lung cancer from the death report. 2) The lung cancer risk is 1 if pathology confirmed lung cancer within $n$ years and the follow-up years are larger than or equal to $n$ years. 3) The exam will be excluded otherwise. In this study, we set $n\in [1, 6]$, meaning that the datasets can be used to train a model for predicting lung cancer risk from 1 to 6 years since the current LDCT scan.

In the NLST, the radiology findings were presented in fully structured tables, while in practice radiology reports are usually in free-text or semi-structured format. Thus, we convert all input data elements into free text descriptions in a template, as shown in Figure~\ref{fig:input}, to fit the practical settings. This template-based conversion lacks diversity and flexibility in describing individual risk factors and imaging findings. To address this issue, we prompted an LLM to augment the input text in different styles, including a table format, a free-text format, and a medical doctor-describing format. The complete input text also includes a question/instruction, e.g., "What are the chances of the patient developing lung cancer within four years post-second-year CT scan?" or "Estimate the lung cancer occurrence risk score for a four-year period after the second-year CT scan." Each question was randomly generated by an LLM given the time information. Then, the responses were prepared corresponding to the ground-truth labels and the training paradigms.

\subsection{Supervised Fine-tuning and Distillation}
Supervised fine-tuning (SFT) is a crucial step in adapting LLMs to perform specific tasks and align better with human expectations. In this study, SFT specializes a general-purpose model for lung cancer risk assessment by further training an LLM on labeled datasets, where each prompt is paired with a target response. However, in lung cancer risk assessment, it is too expensive to obtain large-scale labeled responses for the whole process of calculating the risk score as described in subsection 2.1.
To overcome this challenge, we leverage the distillation technique to get such responses with a large general-purpose LLM. With rapid evolution, open-source LLMs are closing the gap relative to the proprietary models, providing great opportunities for the development of advanced models in specialty domains. In this study, we locally deployed the state-of-the-art open-source LLM as the teacher model without privacy concerns, and a smaller LLM as the student model with limited computational resources. A rejection sampling algorithm was implemented to filter out false responses based on the predicted risk score and the ground truth. We have explored both plain and thinking modes. 
The loss function is
\begin{equation}
\mathcal{L}_{\text{SFT}}(\theta) = - \frac{1}{N} \sum_{i=1}^{N} \sum_{t=1}^{T_i} 
\log P_\theta \left( \bm{o}^{(i)}_t \mid \bm{x}^{(i)}, \bm{o}^{(i)}_{<t} \right)
\end{equation}
where N is the number of training samples, $\bm{o}$ is the target sequences obtained from the teacher model followed by rejection sampling algorithm, $\bm{o}=\bm{y}^*$ for the plain mode, and $\bm{o}=(\bm{r}^*, \bm{y}^*)$ for the thinking mode, $\bm{r}^*$ and $\bm{y}^*$ are the reasoning and answering responses, respectively.

\subsection{Reinforcement Learning for Lung Cancer Risk Assessment}

In contrast to SFT that relies on supervised data, reinforcement learning has become a core technique to enhance reasoning capabilities of LLMs through long CoT thinking even without any SFT as a cold start. 
Inspired by Group Relative Policy Optimization (GRPO)~\cite{shao2024deepseekmath} and Decouple Clip and Dynamic sAmpling Policy Optimization (DAPO)~\cite{yu2025dapo}, we maximize the following objective function for lung cancer risk (LCR) assessment:

\begin{equation}
\begin{aligned}
\mathcal{J}_{\text{LCR}}(\theta) = &
\mathbb{E}_{(x,a)\sim \mathcal{D}, \{o_i\}_{i=1}^G \sim \pi_{\theta_{\text{old}}}(\cdot|x)} 
\Bigg[ 
\frac{1}{\sum_{i=1}^G |o_i|} 
\sum_{i=1}^G \sum_{t=1}^{|o_i|}
 \min \Bigg( 
\frac{\pi_\theta(o_{i,t} \mid x, o_{i,<t})}{\pi_{\theta_{\text{old}}}(o_{i,t} \mid x, o_{i,<t})}\hat{A}_{i,t}, \\
& \text{clip}\big( \frac{\pi_\theta(o_{i,t} \mid x, o_{i,<t})}{\pi_{\theta_{\text{old}}}(o_{i,t} \mid x, o_{i,<t})}\hat{A}_{i,t},(\theta), 1-\varepsilon_{\text{low}}, 1+\varepsilon_{\text{high}} \big)\hat{A}_{i,t}
\Bigg) 
\Bigg]
\end{aligned}
\end{equation}

\begin{align}
\hat{A}_{i,t} = \frac{R_i - \text{mean}(\{R_i\}_{i=1}^G)}{\text{std}(\{R_i\}_{i=1}^G)},
\end{align}
where $\pi_\theta$ and $\pi_{\theta_{\text{old}}}$ are the current and old policy models, and $x$, $o$ are inputs and outputs sampled from the LCR dataset and the old policy $\pi_{\theta_{\text{old}}}$, respectively. $\epsilon_{\text{low}}$ and $\epsilon_{\text{high}}$ are clipping-related hyper-parameters for stabilizing training~\cite{schulman2017proximal, yu2025dapo}, $\hat{A}_{i,t}$ is the advantage computed using a group of rewards $\{R_1, R_2, \cdots, R_G \}$ corresponding to the outputs within each group, $G$ is the number of generations per group and we set $G=8$ in this study.
In reinforcement learning, the reward functions, as the primary training signal, determine the optimization trajectory of the learning process.
Since we have binary ground-truth labels for lung cancer risk, a rule-based reward system was designed, consisting of three components: score reward, format reward, and length penalty. The score reward function $f_{\text{score}}$ is desinged as
\begin{equation}
\begin{aligned}
    f_{\text{score}}(s,\ell; t_1,t_2)
&= (1-\ell)\,\big[(1-2s) + 2s\,\mathbf{1}\{s\le t_1\}\big]
\;+\;
\ell\,\big[(2s-1) + (2-2s)\,\mathbf{1}\{s> t_2\}\big]\\
&=
\begin{cases}
\text{if }\ell=0:\ \begin{cases}
1, & s \le t_1,\\
1-2s, & s> t_1,
\end{cases}\\[6pt]
\text{if }\ell=1:\ \begin{cases}
2s-1, & s \le t_2,\\
1, & s> t_2.
\end{cases}
\end{cases}
\end{aligned}
\end{equation}
where $s$ is the extracted score from the answer part, $l$ is the binary ground-truth calculated as described in Subsection 2.2, $t_1$ and $t_2$ are hyperparameters, we empirically set $t_1=0.45$ and $t_2=0.55$. It is worth mentioning that reward hacking was observed when we simply set $t_1 = t_2 = 0.5$. When $s$ cannot be parsed, the reward is -1.
The format reward function is defined as
\begin{equation}
\begin{aligned}
    f_{\text{format}}(\text{text}) = &\mathbf{1}(\text{text} = <\text{think}> * <\text{/think}>*) \\
    &+ 0.5 \  \mathbf{1}(<\text{think}> \in \text{text}) + 0.5 \  \mathbf{1}(<\text{/think}> \in \text{text})
\end{aligned}
\end{equation}
where $\mathbf{1}$ is an indicator function. To control computational cost, the length penalty function is designed as
\begin{equation}
f_{\text{length}} =
\begin{cases}
0, & l < l_{\max}, \\[6pt]
-\cos\!\left( \dfrac{(l - l_{\max})}{\,l_{\text{completion}} - l_{\max}\,} \cdot \dfrac{\pi}{2} \right), & l \geq l_{\max},
\end{cases}
\end{equation}
where $l_{\text{completion}}$ is the maximum length of the model output, and $l_\text{max}$ is a hyperparameter, above which a cosine penalty is implemented. We empirically set $l_{\text{completion}}=10,000$ and $l_{\text{max}}=9,000$. The final reward function is
\begin{equation}
    f_{\text{reward}} = \alpha f_{\text{score}} + \beta f_{\text{format}} + f_{\text{length}}
\end{equation}
where we simply set $\alpha=\beta=1$.

\section{Results}\label{sec3}

\begin{table}[h!]
\centering
\caption{Constructed datasets. The number of samples represents unique samples.}
\begin{tabular}{m{1\linewidth}}
    \includegraphics[width=\linewidth,page=1]{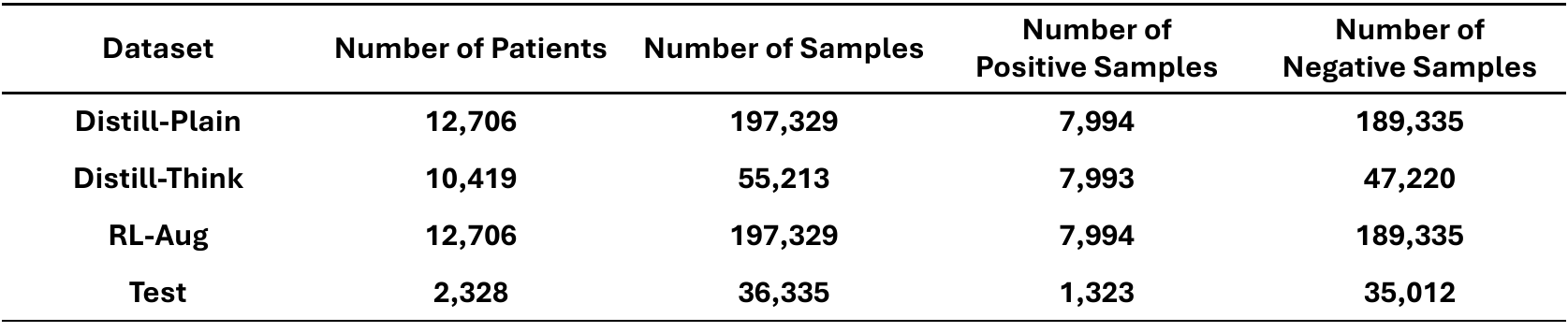}
\end{tabular}
\label{tab:dataset}
\end{table}

\begin{figure}
    \centering
    \includegraphics[width=0.9\linewidth]{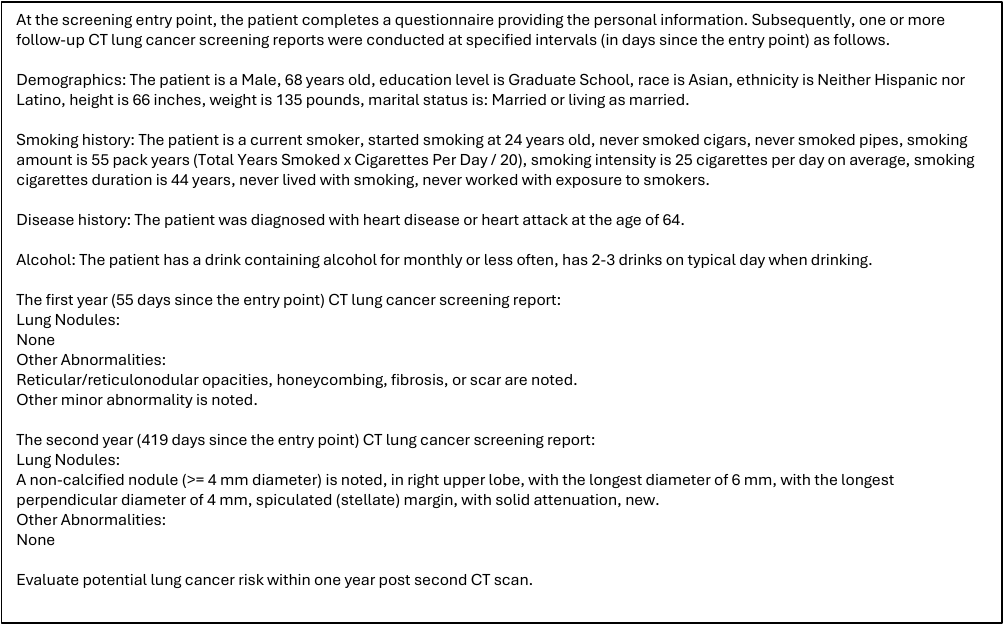}
    \caption{An example of model input.}
    \label{fig:input}
\end{figure}

\subsection{Datasets}
In our experiments, we used Qwen3-235B-A22B as the teacher model to generate the responses for SFT via prompt engineering. In the plain mode, we prompted the model with both inputs and ground-truth labels, and let the model give the risk calculation process with the correct answer without mentioning the ground-truth label is known. Then, we only use the answer part without the thinking content. In this way, we found that all samples got correct responses. In the thinking mode, we just input the personal information and prompted the model to generate the responses as defined in Subsection 2.1.
Each response was then verified by extracting its boxed score. If the score is larger than 0.5, then it is a positive prediction; otherwise, it is a negative prediction. By comparing with the ground-truth, all samples with wrong predictions will be filtered out.

We curated four datasets from NLST to support distillation, RL, and final evaluation, as summarized in Table~\ref{tab:dataset}. The Distill-Plain and Distill-Think sets pair each input with teacher responses in the plain-answer and CoT modes, respectively, after rejection sampling against ground-truth labels. RL contains the same scale of inputs and labels but omits teacher responses, serving as the environment for policy optimization. The Test dataset, containing 36,335 samples from 2,328 patients, is held out at the patient level to prevent leakage across splits. Reflecting the clinical incidence of near-term cancer in screening
cohorts, the positive class prevalence is low ($\sim$4–5\%). These datasets jointly enable: (i) supervised specialization, (ii) reasoning-aware distillation, (iii) reward-driven policy learning with formatting/length control, and (iv) unbiased performance estimation on unseen patients. 

\subsection{Key Results}

\begin{figure}
    \centering
    \includegraphics[width=1\linewidth]{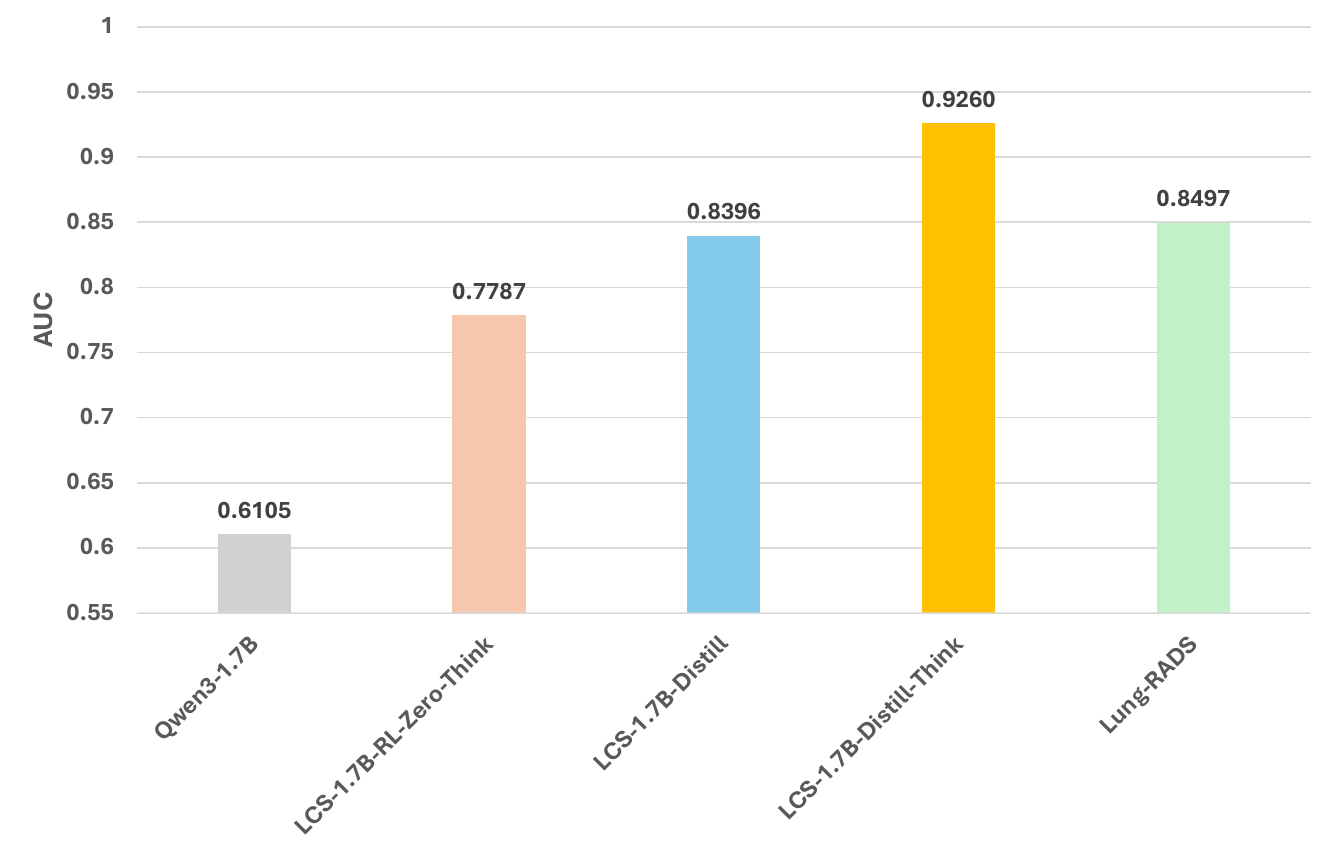}
    \caption{Initial results for 1-year lung cancer risk prediction. The AUC results are reported.}
    \label{fig:results}
\end{figure}

\begin{table}[h!]
\centering
\caption{AUC Results for Multi-Year Risk Prediction.}
\begin{tabular}{m{1\linewidth}}
    \includegraphics[width=\linewidth,page=1]{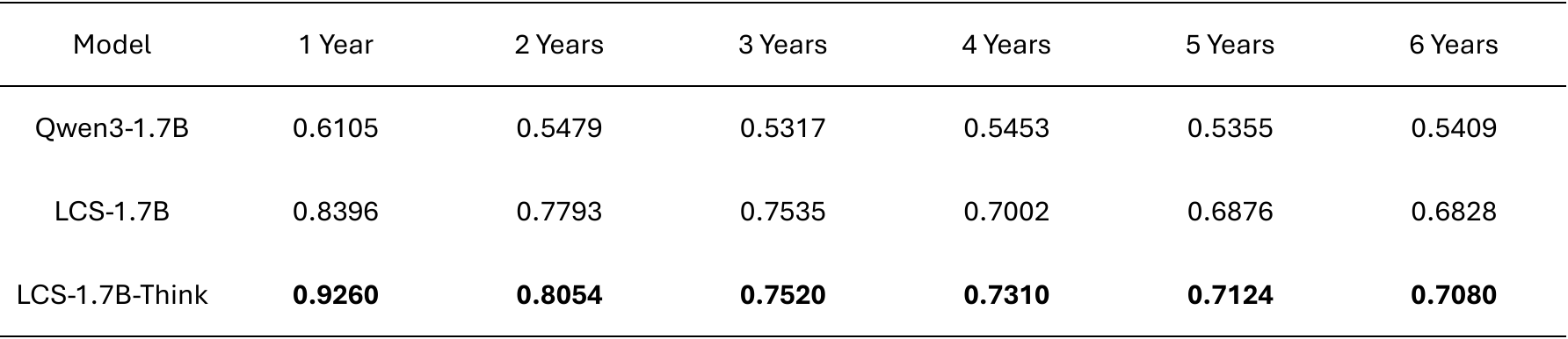}
\end{tabular}
\label{tab:results}
\end{table}

We evaluated multi-horizon risk prediction (1–6 years) using area under the ROC curve (AUC), as shown in Figure~\ref{fig:results} and Table~\ref{tab:results}. The key results are summarized as follows.

\noindent \textbf{Baseline performance:}
The general-purpose Qwen3-1.7B model achieved limited predictive capability (AUC ~0.54–0.61 across years), reflecting the gap between generic LLMs and domain-specific tasks.

\noindent \textbf{Supervised adaptation:}
Our distilled model (LCS-1.7B-Distill) achieved substantial improvement, especially in short-term (1–2 year) predictions, underscoring the value of leveraging teacher-guided reasoning traces.

\noindent \textbf{Reasoning enhancement:}
Explicit reasoning in the thinking mode (LCS-1.7B-Think) further boosted performance. For 1-year prediction, the AUC increased to 0.926, significantly outperforming both Qwen3-1.7B and the plain distilled variant. This highlights the benefit of encouraging structured reasoning in lung cancer risk assessment.

\noindent \textbf{Comparison to Lung-RADS:}
Importantly, all reasoning-enabled models consistently outperformed Lung-RADS across prediction horizons, particularly in short-term risk estimation where timely, actionable decisions are critical.

\noindent \textbf{Reinforcement learning:}
LCS-1.7B-RL-Zero-Think improves substantially over the general baseline (1-year AUC 0.7787 vs. 0.6105), yet trails the distilled models from much larger-size teacher models, suggesting that carefully filtered strong teacher trajectories and answer formats provide richer signals than scalar rewards alone for training a light-weight LLM.

\begin{figure}
    \centering
    \includegraphics[width=0.9\linewidth]{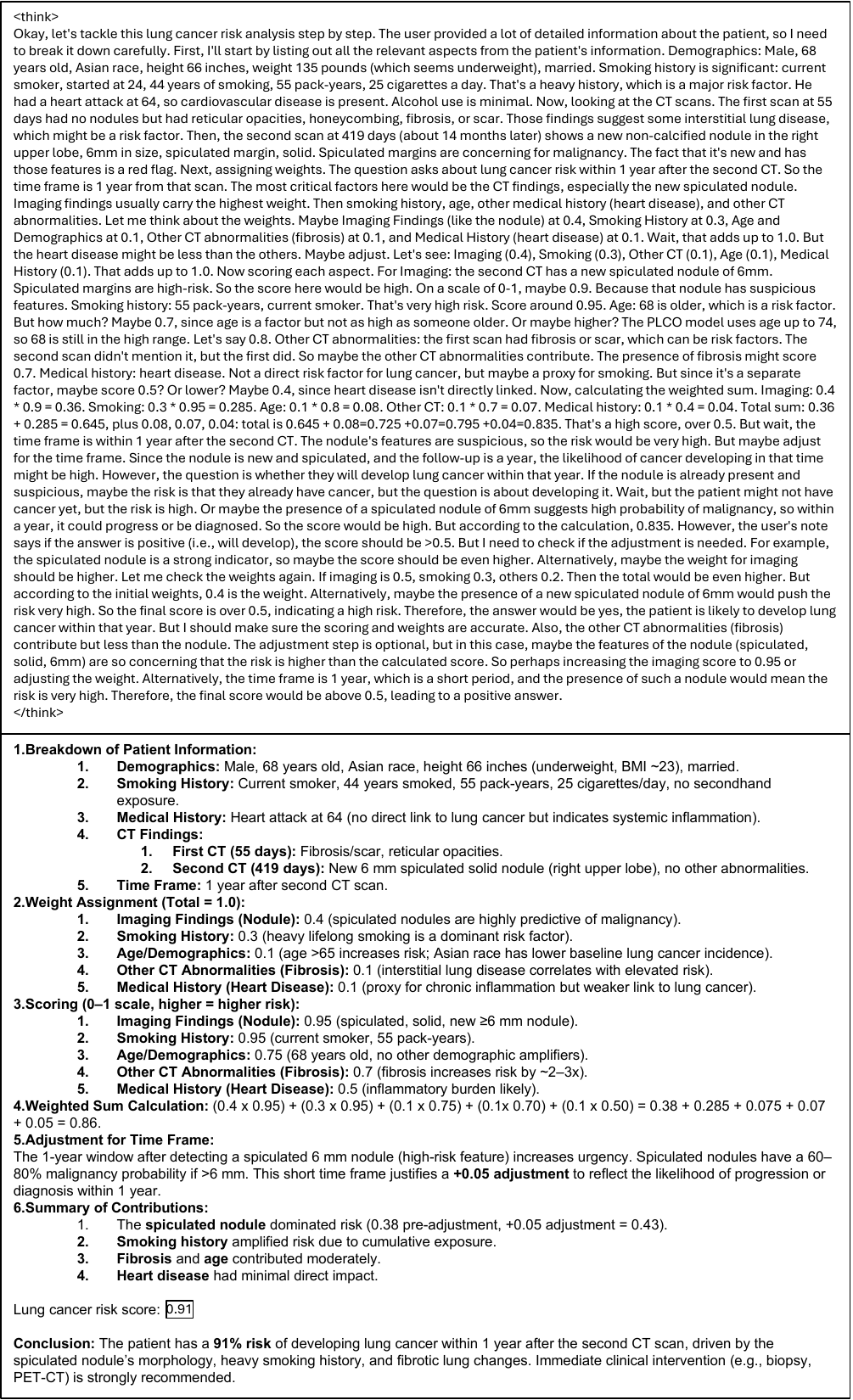}
    \caption{An example of reasoning LLM output corresponding to the input in Figure~\ref{fig:input}.}
    \label{fig:response}
\end{figure}

\noindent \textbf{Monitorability promises:}
CoT outputs reveal the factor-by-factor reasoning process (e.g., weighting a new spiculated 6 mm nodule, heavy smoking, and interstitial changes), culminating in a transparent data-driven equation for lung cancer risk estimation, as shown in Figures~\ref{fig:input} and \ref{fig:response}). This enables targeted inspection of failure modes (e.g., over-weighting a factor) during both training and review by checking for consistency and factual errors in the reasoning process.

\section{Discussions}
This work has demonstrated that a reasoning-enabled LLM can integrate longitudinal LDCT findings with individualized risk factors to produce both accurate and monitorable lung cancer risk predictions. Three aspects are particularly noteworthy further investigation.

Reasoning promotes structured discovery and extracts critical information. Distilled CoT improves performance across horizons, with the largest gain for the 1-year prediction when clinical decision is most time-sensitive. The explicit breakdown of risk contributors (e.g., nodule morphology/trajectory, cumulative smoking exposure, age, and co-morbid lung disease) acts as a structured inductive bias. In contrast to pure end-to-end detectors, the model’s intermediate textual rationales make it feasible to spot miss-weighting, improper context, or contradictions, facilitating iterative correction during human-AI interaction for development and deployment.

Distillation and RL are complementary. Distillation from a strong teacher supplies high-fidelity trajectories and consistent formatting, while RL sharpens policy behaviors directly against task rewards (answer correctness, format compliance, and compute-aware length). In our setting, distillation yielded the highest AUCs, whereas RL provided meaningful improvements over a general LLM but did not match distillation, likely reflecting the richer supervision signal in verified CoT traces versus sparse rewards for training a light-weight LLM. If computational resources are not limited, RL would lead to better results by directly training a large-size LLM when no stronger teacher model is available. Note that the reward design required care (e.g., asymmetric thresholds $t_1=0.45$, $t_2=0.55$) to mitigate reward hacking around the 0.5 decision boundary. 

Clinical advantages of our RLM are substantial over Lung-RADS. The 1-year AUC improvement over Lung-RADS suggests added value from combining imaging features with individualized factors in a longitudinal context. Clinical translation will require mapping continuous risk outputs to management actions (e.g., interval imaging, PET-CT, biopsy) with thresholds tuned to site-specific prevalence and resource constraints, and prospective evaluation of downstream outcomes (e.g., invasive procedure rates, stage shift, and mortality).

There are several limitations for current study: (1) Cohort and generalizability: NLST eligibility and era-specific practices may limit external validity; multi-institutional and contemporary validations are needed. (2) Report synthesis: We converted structured NLST findings into free-text templates and augmentations. While this improves variety, it may not capture the full linguistic variability of real-world reports. (3) Reward design and robustness: Scalar rewards are susceptible to specification gaming. Future work should incorporate multi-dimensional outcome-aware rewards and adversarial evaluation. (4) Calibration and thresholds: We focused on AUC, but clinical deployment demands well-calibrated probabilities with validated operating points across subgroups.

In our follow-up study, We plan to train and evaluate on external datasets with diverse reporting styles, and integrate raw image features (e.g., vision-language encoders) alongside reports to strengthen trajectory-aware nodule assessment. Also, we will add subgroup-aware calibration, and couple CoT with tools (e.g., equation checkers, guideline look-ups) for self-verification. Furthermore, we should conduct prospective reader-in-the-loop studies to improve accuracy, robustness, trust, time-to-decision, and safety, and regulatory compliance. 

\bibliography{sn-bibliography}% common bib file

\end{document}